\documentclass[sigconf]{acmart}

\usepackage{amsmath} % for \boldsymbol macro
\usepackage{bm}      % for \bm macro
\usepackage{multirow} % for \multirow
\usepackage{balance}
\AtBeginDocument{%
  \providecommand\BibTeX{{%
    \normalfont B\kern-0.5em{\scshape i\kern-0.25em b}\kern-0.8em\TeX}}}

\copyrightyear{2023} 
\acmYear{2023} 
\setcopyright{acmlicensed}
\acmConference[ICMI '23]{INTERNATIONAL CONFERENCE ON MULTIMODAL INTERACTION}{October 9--13, 2023}{Paris, France}
\acmBooktitle{INTERNATIONAL CONFERENCE ON MULTIMODAL INTERACTION (ICMI '23), October 9--13, 2023, Paris, France}
\acmPrice{15.00}
\acmDOI{10.1145/3577190.3614114}
\acmISBN{979-8-4007-0055-2/23/10}%%
\begin{document}

\title{Component attention network for multimodal dance improvisation recognition}

\author{Jia Fu}
\orcid{0009-0004-3798-8603}
\affiliation{%
  \institution{RISE Research Institutes of Sweden\\KTH Royal Institute of Technology}
  \city{Stockholm}
  \country{Sweden}
}

\author{Jiarui Tan}
\orcid{0009-0004-0768-5505}
\affiliation{%
  \institution{KTH Royal Institute of Technology}
  \city{Stockholm}
  \country{Sweden}}

\author{Wenjie Yin}
\orcid{0000-0002-7189-1336}
\affiliation{%
  \institution{KTH Royal Institute of Technology}
  \city{Stockholm}
  \country{Sweden}
}

\author{Sepideh Pashami}
\orcid{0000-0003-3272-4145}
\affiliation{%
 \institution{RISE Research Institutes of Sweden}
 \city{Stockholm}
 \country{Sweden}}

\author{M{\aa}rten Bj{\"o}rkman}
\orcid{0000-0003-0579-3372}
\affiliation{%
  \institution{KTH Royal Institute of Technology}
  \city{Stockholm}
  \country{Sweden}}

\renewcommand{\shortauthors}{Jia Fu, et al.}

\begin{abstract}
  % A clear and well-documented \LaTeX\ document is presented as an article formatted for publication by ACM in a conference proceedings or journal publication. Based on the ``acmart'' document class, this article presents and explains many of the common variations, as well as many of the formatting elements an author may use in the preparation of the documentation of their work.

Dance improvisation is an active research topic in the arts. Motion analysis of improvised dance can be challenging due to its unique dynamics. Data-driven dance motion analysis, including recognition and generation, is often limited to skeletal data. However, data of other modalities, such as audio, can be recorded and benefit downstream tasks. This paper explores the application and performance of multimodal fusion methods for human motion recognition in the context of dance improvisation. We propose an attention-based model, component attention network (CANet), for multimodal fusion on three levels: 1) feature fusion with CANet, 2) model fusion with CANet and graph convolutional network (GCN), and 3) late fusion with a voting strategy. We conduct thorough experiments to analyze the impact of each modality in different fusion methods and distinguish critical temporal or component features. We show that our proposed model outperforms the two baseline methods, demonstrating its potential for analyzing improvisation in dance.

\end{abstract}






% \received{8 May 2023}
% \received[revised]{-}
% \received[accepted]{-}


\maketitle

\section{Introduction}
\label{sec:intro}

Dance improvisation is an integral part of contemporary dance, allowing dancers to create spontaneous movements in response to external stimuli or internal impulses. Automated systems for recognizing dance improvisation can aid in the analysis of a dancer's originality, skill, and individuality. Such systems can also provide dance educators with a valuable tool for understanding the cognitive and physiological mechanisms underlying choreography and assessing the effectiveness of improvisation training programs. However, despite its importance in the dance world, improvisation has received relatively little attention in the field of computer vision and machine learning. 

Most existing research on dance recognition focuses on the discrimination of dance types \cite{protopapadakis2017folk, wysoczanska2020multimodal, hu2021unsupervised} or the classification of movements in a specific type of dance \cite{hendry2020development, matsuyama2021deep, bhuyan2022motion}. There is a lack of research highlighting the identification of different expressive qualities in improvisational dance. Although frameworks that perform well on general human action prediction tasks \cite{butepage2017deep, yang2020group, win2020real} can also be employed in improvisational dance recognition, they are limited in unimodal skeletal data. Besides skeleton dynamics, data of other modalities, e.g., inertial measurement unit (IMU) signals and respiratory cadence, can change idiosyncratically with dance moves. In order to investigate how to make better use of multimodal information for improvisational dance classification, we carry out this work and make the following contributions:

Firstly, we propose multimodal fusion methods in three levels: 1) feature fusion by a component attention network (CANet) adapted from BodyAttentionNet \cite{wang2019learning}; 2) model fusion by fusing a graph convolutional network (GCN) \cite{welling2016semi} with CANet; 3) and late fusion by a simple voting. Our proposed fusion strategies exceed the two State-of-the-Art frameworks in multimodal human motion prediction. Furthermore, we analyze the temporal/component attention scores and visualize them with heat maps, which leads to a quantifying of creative expression. The source code is available for verification.~\footnote{https://github.com/JasonFu1998/ComponentAttentionNetwork}

\section{Related work}
\textbf{Multimodal fusion} is among the most critical topics in multimodal learning. It aims to aggregate information from multiple modalities to infer discrete labels for classification tasks, such as audio-visual speech recognition \cite{gurban2008dynamic}, or continuous values for regression tasks, such as emotion prediction \cite{baltruvsaitis2013dimensional}. Three paradigms are characterized based on the stages when fusion is conducted: feature-level fusion, decision-level fusion, and model-level fusion. Feature-level fusion (a.k.a. early fusion) concatenates feature vectors right after they are extracted from various single modalities. \cite{busso2004analysis} successfully achieves a feature-level fusion of facial expressions and speech. However, feature-level fusion can be hindered by the high dimensionality of the feature space \cite{zhao2018multi}. Decision-level fusion (a.k.a. late fusion) gives the final decision by incorporating the predictions inferred from different modalities in a voting process. It can be well applied to some multimodal inference tasks, such as affective computing \cite{ringeval2015av+}, by ignoring low correlation modalities.  Model-level fusion integrates intermediate representations of different modalities, which balances the benefits of the two approaches mentioned above and demonstrates effectiveness in \cite{nicolaou2011continuous}.

\noindent \textbf{Dance recognition.} Dance motion data typically involves both temporal and spatial information. Classic approaches, such as long short-term memory (LSTM) \cite{hochreiter1997long}, gated recurrent unit (GRU) \cite{cho2014learning}, and GCN, are common solutions for handling such data. Some studies have attempted to address multimodal data of human movements. Fusion-GCN \cite{duhme2022fusion} incorporates other modalities, e.g., RGB data and IMU signals, into the GCN that represents the human skeleton. Gimme Signals \cite{memmesheimer2020gimme} provides another novel approach for capturing motion features from multimodal data: signals of different modalities but the same length can be plotted in the same image. Afterward, such images can be used to train convolutional neural networks (CNNs). In particular, there has been extensive research exploring the applications of machine learning (ML) and deep learning (DL) in the classification of dance genres or dance figures. \cite{protopapadakis2017folk} compares the performance of DL and traditional ML methods on dance classification with Kinect sensor data. In \cite{matsuyama2021deep}, a neural network is trained to classify 3D pose data from wearable sensors into different ballroom dances. \cite{wysoczanska2020multimodal} designs a late fusion network for multimodal dance classification, involving four different modalities: RGB frames, optical flows, skeletal data, and audio.

\section{Methodology}
% Formulate the problem with equations and symbol
\textbf{Problem Formulation.} We focus on N-class classification on multimodal data. Each sample is a sequence of length $T$. At a time step $t$, each sample consists of $C$ \textit{components}, each of which is a feature vector $\bm{f}_t^{(c)}$. Such vectors of different components may belong to the same or different modalities and therefore are not necessarily of the same size. Our goal is to train a neural network $\phi$ that serves as a mapping from input features $\{\bm{f}^{(1)}_{1:T}, \bm{f}^{(2)}_{1:T}, \dots, \bm{f}^{(C)}_{1:T}\}$ to probabilities of categories $\bm{p} = [p_1, p_2, \dots, p_N]$. The predicted category is given by $n = \arg \max_{i} p_i$. In our scenario, a binary classification task for dance movement data is addressed, where $N=2$.

\begin{figure}
    \centering
    \includegraphics[width=0.95\linewidth]{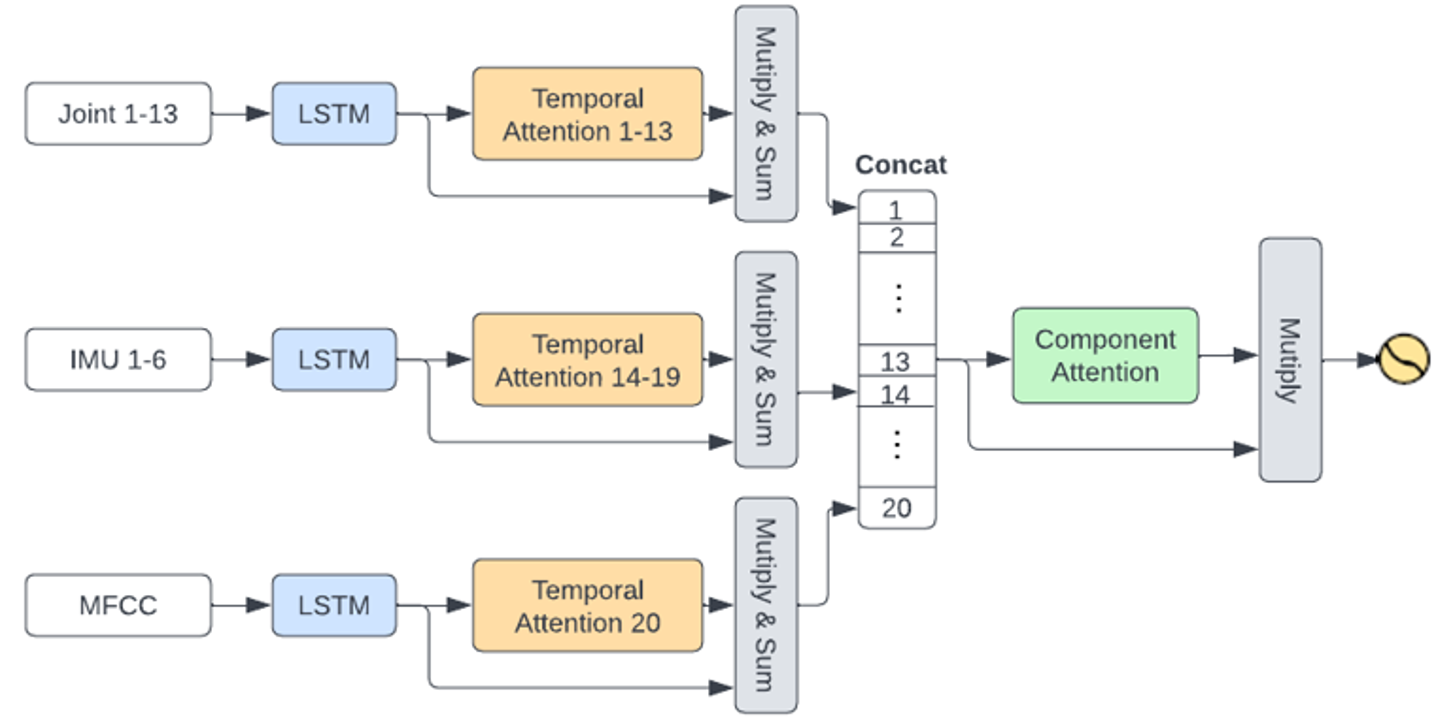}
    \caption{The Network Architecture of CANet. Different input features are fed into their respective branches, each containing a shared LSTM and a temporal attention module. Extracted information from different branches is concatenated and then fused by a component attention module.}
    \Description{This figure shows the network architecture of CANet, whose content is described in the caption and details can be found in the METHODOLOGY section.}
    \label{fig:CANet}
\end{figure}

\noindent \textbf{CANet.} We first present CANet, as depicted in Fig.~\ref{fig:CANet}, which can be seen as an upgraded version of BodyAttentionNet \cite{wang2019learning}. It targets multimodal data rather than just body parts. In addition, although CANet is designed for feature fusion, it can be further used for model fusion. To begin with, an input feature matrix is constructed by aligning components by frame and concatenating them. Next, CANet extracts features through a two-stage attention mechanism. Temporal attention modules extract information separately for different components. A subsequent component attention module performs further information extraction at the global level.

We first analyze each branch of CANet and ignore batch operations. For example, in the branch for component $c$, LSTM outputs a matrix $\bm{H}_{1:T}^{(c)} \in \mathbb{R}^{T \times K}$, where $K$ is the output dimension. The parameters of LSTM are shared by all branches. Temporal attention scores $\bm{a}_{1:T}^{(c)} \in \mathbb{R}^{T}$ are calculated and then normalized by the softmax function: $\bm{a}_{1:T}^{(c)}=\operatorname{Softmax}\left(\bm{H}_{1:T}^{(c)} \bm{w}^{(c)} \right)$,
where $\bm{w}^{(c)} \in \mathbb{R}^{K} $ are trainable parameters. The output of the temporal attention layer $\bm{\theta}^{(c)} \in \mathbb{R}^{K}$ is a weighted sum of information from different time steps: $\bm{\theta}^{(c)} = \bm{H}_{1:T}^{(c) \top} \bm{a}_{1:T}^{(c)}$.

The outputs from temporal attention layers are then concatenated into a matrix $\bm{\Theta} = [\bm{\theta}^{(1)}, \bm{\theta}^{(2)}, \dots, \bm{\theta}^{(C)}] \in \mathbb{R}^{K \times C}$. By this point, the independent branches of the network are merged. Then the component attention module calculates a component attention map $\bm{B} \in \mathbb{R}^{K \times C}$ as follows: $\bm{B}=\operatorname{Softmax} \Bigl( \operatorname{tanh} \left( \bm{\Theta} \bm{W}_1 + \bm{b}_1 \right) \bm{W}_2 + \bm{b}_2 \Bigl)$,
where $\bm{W}_i, \bm{b}_i$ denote the weight and bias of a fully connected layer, the same as below. This attention map reflects the different importance of the components in different modalities and is subsequently used to weight the outputs. Weighted output matrix $\bm{O} \in \mathbb{R}^{K \times C}$ embeds an entire input sample and is given by $\bm{O} = \bm{B} \odot \bm{\Theta}$, where $\odot$ denotes element-wise multiplication. In the end, the predicted probabilities $\bm{p} \in \mathbb{R}^{1 \times N}$ are computed as follows: $\bm{p} = \operatorname{Softmax} \Bigl(\operatorname{vec} (\bm{O}) \bm{W}_3 + \bm{b}_3 \Bigl)$,
where $\operatorname{vec}$ denotes flattening a matrix into a row vector.

\begin{figure}
    \centering
    \includegraphics[width=0.95\linewidth]{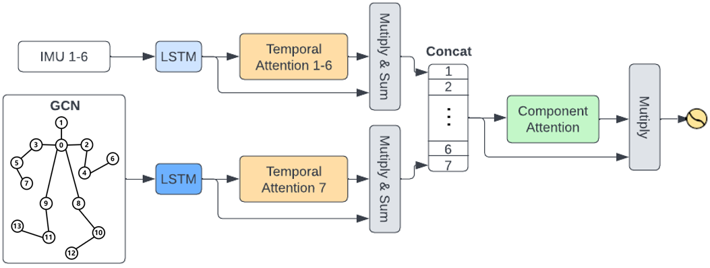}
    \caption{The Structure of GCN-CANet. Its modification over CANet is that the features of different joints are first integrated into one component using a GCN. In addition, the LSTM in this branch is not shared.}
    \Description{This figure shows the structure of GCN-CANet, whose content is described in the caption and details can be found in the METHODOLOGY section.}
    \label{fig:model-fusion}
\end{figure}

\noindent \textbf{GCN-CANet}, as exhibited in Fig.~\ref{fig:model-fusion}, upgrades CANet by constructing an undirected graph of keypoints extracted from the human body. One special branch of GCN-CANet, termed GCN-LSTM in this paper, contains several cascading GCN layers and LSTM layers. It outputs intermediate features $\bm{H}_{1:T}^{(g)} \in \mathbb{R}^{T \times K}$, the dimensionality of which remains unchanged compared to CANet. This branch captures the spatio-temporal dynamics of human skeletons, and the intermediate features can be fused at the component attention module without modifying the main architecture of CANet.
 
\noindent \textbf{Decision voting.} Given predictions from multiple models, the final decision is made by voting, which is regarded as late fusion.

\section{Experiments}

\textbf{Data preparation.} We use the Unige-Maastricht Dance dataset \cite{niewiadomski2019does, vaessen2019computational}. It contains 152 improvisational dance segments of two expressive qualities known as lightness or fragility. The primary criterion for distinguishing them is whether the fluidity is presented without interruption. Each segment is recorded in four modalities: dance video, IMU data, respiration audio, and electromyography (EMG) signals. They are synchronized at 50fps with an average length of 10.3s. 130 segments are randomly chosen to construct the training set, and the remaining 22 segments are used for testing. Based on \cite{wang2019recurrent}, we run a sliding window to produce data instances from each segment, with a length of 3s and an overlap ratio of 80\%.

\begin{figure}
    \centering
    \includegraphics[width=0.95\linewidth]{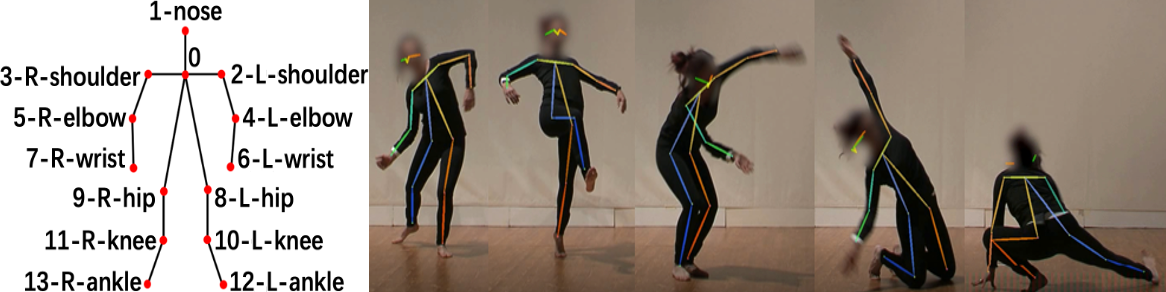}
    \caption{Body Joints Ordering and Extracted Skeletons.}
    \Description{Body graph vertices are: 1.nose, 2.left shoulder, 3.right shoulder, 4.left elbow, 5.right elbow, 6.left wrist, 7.right wrist, 8.left hip, 9.right hip, 10.left knee, 11.right knee, 12.left ankle, and 13.right ankle (keypoint 0 is the midpoint between keypoint 2 and 3). Edges are: 0-1, 0-2, 0-3, 0-8, 0-9, 2-4, 3-5, 4-6, 5-7, 8-10, 9-11, 10-12, and 11-13. This figure also shows the body skeleton as such a graph in five example video frames.}
    \label{fig:skeleton}
\end{figure}

The pose of a dancer in the video is estimated with AlphaPose \cite{fang2017rmpe, li2019crowdpose}, which has been pretrained on COCO 2017 \cite{lin2014microsoft}. AlphaPose estimates the coordinates and visibility $(x, y, v)$ of 17 keypoints for each detected person. We adopt the nose keypoint to mark the position of the dancer's head, with other facial keypoints deprecated for simplicity. Fig.~\ref{fig:skeleton} displays the body graph and the correspondence between ordered keypoints and body joints. The dancer wears IMUs on the left and right hands. Each IMU outputs $(x, y, z)$ of the accelerometer, gyroscope, and magnetometer, respectively. As for audio, we compute the Mel-scale frequency cepstral coefficients (MFCC) \cite{logan2000mel} with 13 dimensions, which is the most common feature in speech recognition. Notice we abandon EMG in experiments due to some segments losing one of two EMG signals. In summary, there are 13 Joints components, 6 IMU components (2 accelerometers, 2 gyroscopes, and 2 magnetometers), and 1 MFCC component.

In our implementation of CANet and GCN-CANet, each LSTM has 3 layers with 8 hidden units, and each GCN consists of 3 layers with 16 hidden units. Models were trained with Adam optimizer \cite{kingma2014adam}. We first conducted experiments on the individual modalities, followed by incorporating ancillary modalities, IMU and MFCC, into Joints modality through three fusion methods. Table \ref{ours-fusion} summarizes the performance comparison.

\begin{table}
\begin{center}
\caption{Results of Three Fusion Strategies. For late fusion, the best models (GCN-CANet, CANet, GRU) for corresponding modalities (J: Joints, I: IMU, M: MFCC) are used for voting.}

\begin{tabular}{ccccc}
  \toprule
  Fusion Level & Modalities & Model & Accuracy & F1 \\
  \midrule
  - & J & CANet & 74.26\% & 0.74 \\
  - & \textbf{J} & \textbf{GCN-CANet} & \textbf{76.79\%} & \textbf{0.77}  \\
  - & \textbf{I} & \textbf{CANet} & \textbf{76.37\%} & \textbf{0.76}  \\
  - & M & CANet & 56.54\% & 0.54  \\
  - & \textbf{M} & \textbf{GRU} & \textbf{59.49\%} & \textbf{0.59}  \\
  \textbf{Feature} & \textbf{J + I} & \textbf{CANet} & \textbf{82.28\%} & \textbf{0.82}  \\
  Feature & J + M & CANet & 68.78\% & 0.68 \\
  Feature & J + I + M & CANet & 77.64\% & 0.78  \\
  \textbf{Model} & \textbf{J + I} & \textbf{GCN-CANet} & \textbf{83.12\%} & \textbf{0.83}  \\
  Model & J + I + M & GCN-CANet & 81.86\% & 0.82  \\
  \textbf{Late} & \textbf{J + I + M} & \textbf{Vote} & \textbf{78.90\%} & \textbf{0.79}  \\
  \bottomrule
\end{tabular}

\label{ours-fusion}
\end{center}
\end{table}

\noindent \textbf{Feature fusion.} We can conclude that: 1) For unimodal in CANet, IMU works best, and MFCC performs only marginally better than the random classification. 2) Joints and IMU are fused effectively. 3) Fusing MFCC will degrade the results compared to excluding it. 

\begin{figure*}[htbp]
    \centering
    \includegraphics[width=1\textwidth]{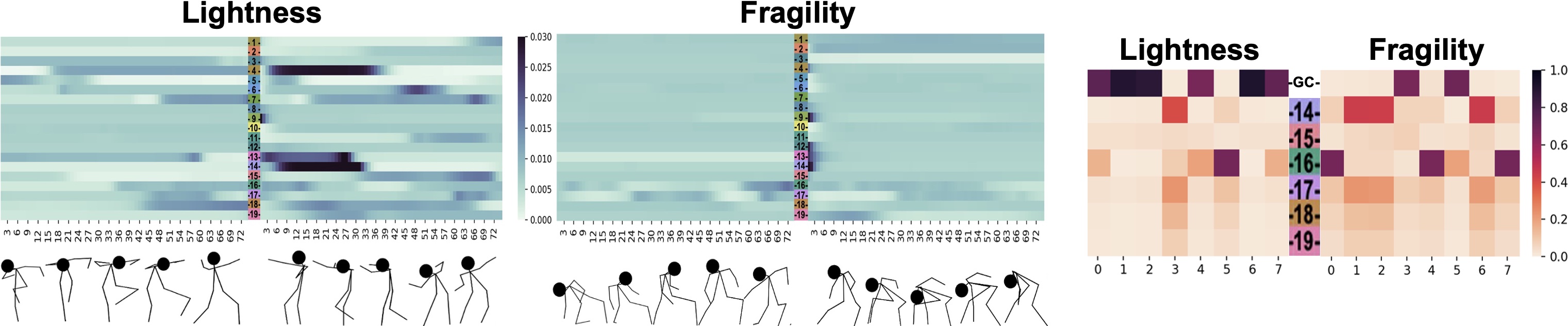}
    \caption{Heat Maps of Temporal (left) and Component (right) Attention Score for Selected Instances. The x-axis are the frame index and the numerical order of the entries in attention score vector respectively, and the y-axis are the index of components which has the same order and color of components listed in Fig. \ref{fig:box}. GC represents graph convolution of the Joints 1 - 13.}
    \Description{The left part of this figure shows temporal attention heat maps for selected two test lightness instances and two test fragility instances. The right part of this figure shows component attention heat maps for selected one test lightness instance and one test fragility instance. More descriptions are available in the caption and DISCUSSION section.}
    \label{fig:heatmap-both}
\end{figure*}

\begin{table}
\begin{center}
\caption{Comparison of Methods on Classification Accuracy.}
\begin{tabular}{cccc}
\toprule
Modalities          &  Ours  & Fusion-GCN \cite{duhme2022fusion}   & Gimme \cite{memmesheimer2020gimme}  \\ 
\midrule
Joints              & 76.79\%          & 76.79\% & 71.39\% \\ 
Joints + IMU        & \textbf{83.12\%} & 73.52\% & 77.67\% \\ 
Joints + IMU + MFCC & \textbf{81.86\%} & 72.25\% &   -     \\
\bottomrule
\end{tabular}
\label{tab:baselines}
\end{center}
\end{table}

\noindent \textbf{Model fusion.} We arrive at the following conclusions from our experiments: 1) GCN-CANet beats plain CANet when they have the same training data, i.e., model-level strategy is preferable to feature-level for the accordant fused components; 2) and linking discrete body joints into an integrated graph component by GCN mitigates the detrimental effect of bringing MFCC to the fusion.

\begin{figure}
    \centering
    \includegraphics[width=0.95\linewidth]{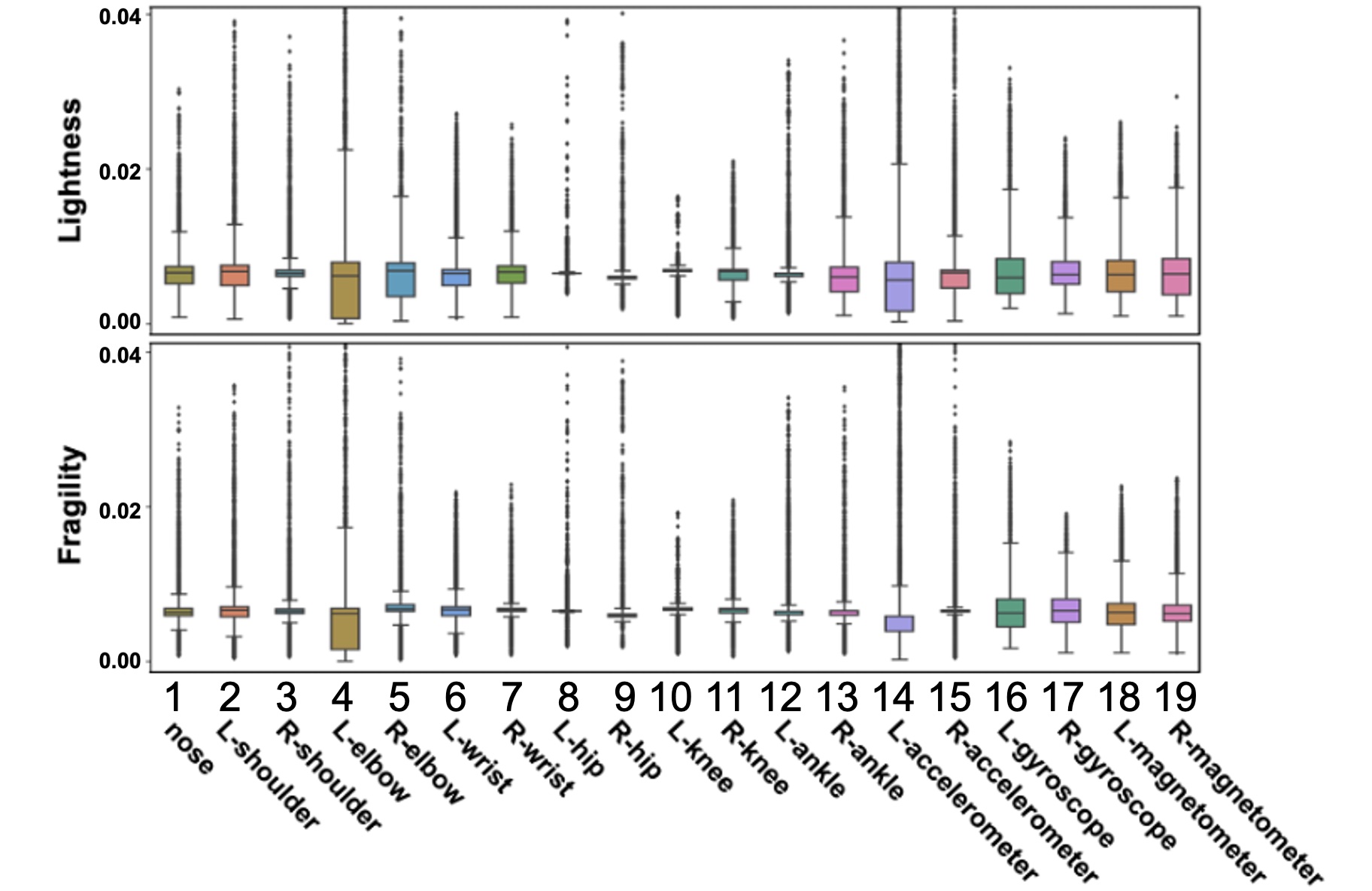}
    \caption{Temporal Attention Scores of All Test Instances. The x-axis lists a total of 19 Joints and IMU components.}
    \Description{This figure shows the boxplots of temporal attention statistics for all test instances predicted and categorized as lightness and fragility. More descriptions are available in the caption and DISCUSSION section.}
    \label{fig:box}
\end{figure}

\noindent \textbf{Decision fusion.} We choose the best-performing model for each modality respectively. Late fusion demands at least triple predictions for the same sample. Unlike before, MFCC is necessary and beneficial for late fusion. With all three modalities, late fusion performs better than feature fusion but worse than model fusion.

\noindent \textbf{Comparison to baselines.} In order to compare our approach with recent works, we conducted three experiments (unimodal, bimodal, and trimodal classification) to compare our methods with two baselines, Fusion-GCN \cite{duhme2022fusion} and Gimme Signals \cite{memmesheimer2020gimme}. In Fusion-GCN, the GCN was implemented the same as ours for a fair comparison, but IMU and MFCC information were fused according to the original approach. The results presented in Table \ref{tab:baselines} illustrate that our approach achieves higher classification accuracy compared to the baselines in all experiments. The fusion scheme of Fusion-GCN resulted in lower accuracy with more modalities, demonstrating that it is not a good choice for this dataset. For Gimme, it is hard to encode data with different dimensions into a single image, especially when some modalities have high dimensions. Therefore, it is not suitable for trimodal data.

\section{Discussion}
\textbf{The deficiency of MFCC.} From the video recordings, it can be observed that fragility movements generally have a narrower spatial range than lightness ones. Some typical fragility situations are performed at a fixed site on the stage throughout the whole segment. Intuitively, smaller motion amplitude and velocity will lead to a smoother breath rhythm. MFCC is initially envisioned to reflect such tempo fluctuations. However, because the interval between two consecutive deep breaths is usually longer than the length of a sliding window, clear breaths captured by the microphone do not appear in all audio instances. Stability and complementarity of the information from each modality are prerequisites for successful fusion, which MFCC falls short of delivering.

\noindent \textbf{Analysis of temporal attention scores.} Based on the best results on CANet, when 13 Joints components and 6 IMU components are fused, we assess the temporal attention scores. Fig.~\ref{fig:box} shows boxplots of attention statistics for all test instances categorized by the predicted motion labels. Boxplots associated with lightness are usually wider than those with fragility for the identical component, especially for left shoulder, right elbow, right ankle, and two accelerometers. This can be traced to two roots.

Dancers show more various choreography to convey lightness than fragility. This may rely on the diverse considerations made by dancers over which body parts should be highlighted as light and suspending. We choose two test lightness instances that are correctly classified with high confidence and then visualize their temporal attention scores in Fig.~\ref{fig:heatmap-both}. For conciseness, only 75 successive frames are clipped from the selected instances. The corresponding dancer sketches below are exhibited around every 15 frames. An attention switch from right elbow to left elbow is evident in the left instance as the dancer's left arm expands gradually from inside to outside of the trunk. More attention is paid on right ankle in the first 60 frames, meanwhile, the dancer's right foot moves from hovering to landing. In the first half of the right instance, the most decisive cues are discovered on left elbow and accelerometer of the left hand. The simultaneous dancer's left arm is stuck still, conforming to the characteristics of lightness. Moreover, the concern for right ankle is also reflected on the attention heat map, where the dancer elevates right foot to execute a horizontal side turn.

Conversely, the limited width of fragility boxplots illuminates the pattern that each joint preserves equally consistent relevance throughout the dance, emphasizing the collaboration of different body parts in maintaining balance on the verge of falling. Fig.~\ref{fig:heatmap-both} contains two representative fragility instances. Almost all components receive even attention throughout, except for gyroscope pairs.

\noindent \textbf{Analysis on component attention scores.} Fig.~\ref{fig:heatmap-both} includes two component attention heat maps in GCN-CANet without MFCC, matching two test instances with high prediction scores on their ground truth. The attention vector for each component rests with the hidden units of LSTM. Graph convolution is given the most attention among the involved components. Additionally, information is held in a complementary manner by different units of the same component. Take the graph convolution as an example, the nature of fragility is coupled with unit 3 and 5, whilst other units are more oriented toward lightness. As for IMU pairs, especially the gyroscope, more effect is carried by the left sensor than the right sensor for both motion qualities, confirming that the dancer's control of the left and the right arms is asymmetrical.

\section{Conclusion}

We employ three multimodal schemes, namely feature fusion, model fusion, and decision fusion, to classify the qualities of dance improvisation. We present CANet for feature fusion, which leverages the attention mechanism to devote more focus to the essential moments and components. By connecting all discrete joints into a human body topology, GCN-CANet for model fusion surpasses the naive CANet and also alleviates the negative effects of incorporating MFCC in the fusion. The experimental and visualized results demonstrate the effectiveness and scientificity of our models, which outperform the baselines in improvisational dance recognition.

\bibliographystyle{ACM-Reference-Format}
\balance
\bibliography{sample-base}

\appendix









\end{document}